%% file: main.tex
\documentclass[pdflatex,sn-mathphys-num]{sn-jnl}

\usepackage{graphicx}%
\usepackage{multirow}%
\usepackage{amsmath,amssymb,amsfonts}%
\usepackage{amsthm}%
\usepackage{mathrsfs}%
\usepackage[title]{appendix}%
\usepackage{xcolor}%
\usepackage{textcomp}%
\usepackage{manyfoot}%
\usepackage{booktabs}%
\usepackage{algorithm}%
\usepackage{algorithmicx}%
\usepackage{algpseudocode}%
\usepackage{listings}%

\theoremstyle{thmstyleone}%
\theoremstyle{thmstyletwo}%

\theoremstyle{thmstylethree}%

\input{defs}

\begin{document}

\title[Article Title]{Directed Social Regard: Surfacing Targeted Advocacy, Opposition, Aid, Harms, and Victimization in Online Media}


\author*[1]{\fnm{Scott} \sur{Friedman}}\email{friedman@sift.net}

\author[1]{\fnm{Ruta} \sur{Wheelock}}\email{rwheelock@sift.net}

\author[2]{\fnm{Sonja} \sur{Schmer-Galunder}}\email{s.schmergalunder@ufl.edu}

\author[1]{\fnm{Drisana} \sur{Iverson}}\email{dmosaphir@sift.net}

\author[1]{\fnm{Jake} \sur{Vasilakes}}\email{jvasilakes@sift.net}

\author[3]{\fnm{Joan} \sur{Zheng}}\email{joan.zheng@cgu.edu}

\author[1]{\fnm{Jeffrey} \sur{Rye}}\email{rye@sift.net}

\author[4]{\fnm{Vasanth} \sur{Sarathy}}\email{vasanth.sarathy@tufts.edu}

\author[1]{\fnm{Christopher} \sur{Miller}}\email{cmiller@sift.net}

\affil*[1]{\orgname{SIFT}, \orgaddress{\street{319 N 1st Ave}, \city{Minneapolis}, \postcode{55401}, \state{MN}, \country{USA}}}

\affil[2]{\orgdiv{Department of Computer \& Information Science \& Engineering}, \orgname{University of Florida}, \orgaddress{\street{432 Newell Drive}, \city{Gainesville}, \postcode{32611}, \state{FL}, \country{USA}}}

\affil[3]{\orgdiv{Center for Information Systems \& Technology}, \orgname{Claremont Graduate University}, \orgaddress{\street{150 E 10th St.}, \city{Claremont}, \postcode{91711}, \state{CA}, \country{USA}}}

\affil[4]{\orgdiv{Department of Computer Science}, \orgname{Tufts University}, \orgaddress{\street{177 College Avenue}, \city{Medford}, \postcode{02155}, \state{MA}, \country{USA}}}

\input{abstract}




\maketitle


\input{intro}

\input{related}

\input{approach}

\input{results}

\input{discussion}


\section*{Conflicts of Interest}
On behalf of all authors, the corresponding author states that there is no conflict of interest.

\section*{Data Availability Statement}
The online media used for annotation, training, and validation (\tabref{dsrdata}) is comprised of publicly accessible information (PAI) and URLs are listed for the various open sources in \tabref{dsrdata}.
The data used for our application studies are available via the MFTC data repository (https://osf.io/k5n7y/), the Ribeiro et al data repository (https://www.doi.org/10.5281/zenodo.4007913), and the Bridging Comments Benchmark Dataset repository (https://github.com/conversationai/Bridging-Comments-Benchmark-Dataset).

Training data containing the specific human annotations of span-level sentiment are unable to be shared due to DoD SBIR rights protections.
The annotation process and user interface for gathering all of these human ratings are illustrated in \secref{annotation} and \figref{slider}.

\bibliography{dsr,related-work-refs}

\input{appendix}

\end{document}

%% file: defs.tex
\newcommand{\textq}[1]{\textit{``#1''}}

\newcommand{\secref}[1]{Section~\ref{sec:#1}}
\newcommand{\tabref}[1]{Table~\ref{tab:#1}}
\newcommand{\figref}[1]{Figure~\ref{fig:#1}}

\newcommand{\tb}[1]{\textbf{#1}}

%% file: abstract.tex
\abstract{
The language in online platforms, influence operations, and political rhetoric frequently directs a mix of pro-social sentiment (e.g., advocacy, helpfulness, compassion) and anti-social sentiment (e.g., threats, opposition, blame) at different topics, all in the same message.
While many natural language processing (NLP) tools classify or score a text's overall sentiment as positive, neutral, or negative, these tools cannot report that positive and negative sentiments coexist, and they cannot report the target of those sentiments.
This paper presents the Directed Social Regard (DSR) approach to multi-dimensional, multi-valence sentiment analysis, comprised of a pair of transformer-based models that (1) detects span-level targets of sentiment in a message and then (2) scores all spans within the message context along three (-1, 1) axes of regard that are motivated by social science theories of moral disengagement and moral framing.
We present a data collection and annotation strategy for DSR dataset construction, a transformer-based architecture for span-level scoring, and a validation study with promising results.
We apply the validated DSR model on six third-party datasets of online media and report meaningful correlations between DSR outputs and the labels and topics in these pre-existing social science datasets.
}

%% file: intro.tex
\section{Introduction}\label{sec1}

Online media can express a broad array of influential sentiments, such as ascribing advocacy, victimization, blame, and opposition toward various topics, and sometimes all in a single sentence.
The sentiment and moral commitments in online messaging have implications for a message's continued existence and viral spread across social networks, potentially propagating outrage across online communities \cite{brady2023overperception}.
Consequently, researchers and online platforms have recently begun measuring multiple types of emotion \cite{guo2022emotion} and sentiment to identify and prioritize prosocial media that has the potential to bridge individuals and communities \cite{schmer2024annotator}.

In addition to understanding the \textit{general} sentiment of media, recent sentiment-analysis approaches are also answering the questions of \textq{sentiment about what?} or \textq{sentiment toward whom?} \cite{cai2024joint,hossain2025dynamic,ronningstad2024entity}.
These questions are important for complex messaging that combines multiple themes---such as advocating that we aid a victimized population and sanction their aggressor---all in the same message.
These within-message complexities are important considerations, and ascribing a single score or category for an entire message may gloss over important pro- or anti-social themes that are \textit{directed} toward different topics in the same context.
Understanding how sentiment is directed at specific actors and topics---such as people, organizations, ideologies, geopolitical entities, products, and events---can inform social science, political analysis, cultural anthropology, market research, and information operations.

To address this gap, we introduce Directed Social Regard (DSR), a novel framework for targeted sentiment analysis built upon theories of moral disengagement in the social sciences \cite{bandura2002selective,bandura1996mechanisms}. DSR refines the task of targeted sentiment analysis to include \textit{dimensions} of regard---interpretable, bipolar axes along which sentiment varies continuously---allowing for richer representations. To support our research, we also introduce a new dataset of DSR annotations over social media data.


\begin{figure}[h]
\centering
\includegraphics[width=.8\textwidth]{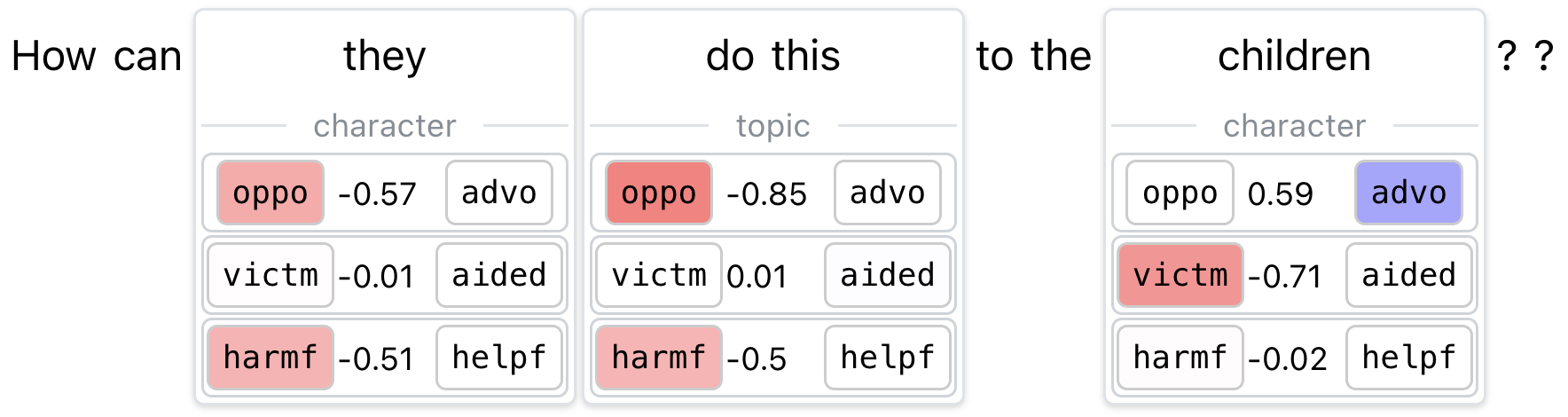}
\caption{Directed Social Regard output for a simple sentence, inferring that the author expresses opposition (\textbf{oppo}, red) and harmfulness (\textbf{harmf}, red) directed at \textq{they} and \textq{do this}, and expresses advocacy (\textbf{advo}, blue) and victimization (\textbf{victm}, red) directed at \textq{children.}}
\label{fig:dsr-ui}
\end{figure}

We illustrate the richness of DSR representations in \figref{dsr-ui}. The input sentence \textq{How can they do this to the children??} exemplifies multi-sentiment messaging that directs a mixture of positive and negative regard at different entities and expresses layered regard toward single entities (e.g., advocating for the children while also expressing their victimization). Our model first identifies character (``they'' and ``children'') and topic (``do this'') spans in the input text. It then scores each span along three bipolar dimensions of regard from the author's perspective: Opposed--Advocated, Victimized--Aided, and Harmful--Helpful.


This paper is structured as follows. We provide an in-depth review of previous work alongside comparisons of DSR to existing sentiment analysis approaches in \secref{related}. \secref{approach} describes DSR's underpinnings in the social sciences and moral disengagement theory before laying out its realization in a transformer-based NLP architecture. We also describe the data collection and annotation process for our new DSR dataset. \secref{results} details experimental results of the span recognition and DSR scoring models on this dataset, followed by example analyses of existing datasets using our DSR model. We close with a discussion of the social science implications and directions for future work.


%% file: related.tex
\section{Related Work}
\label{sec:related}

The task of measuring how language communicates attitudes toward specific entities has been approached under several related paradigms in NLP, including targeted sentiment analysis, stance detection, and aspect-based sentiment analysis. While these frameworks share common goals, they often use overlapping terminology, which can obscure important distinctions. In this section, we clarify our terminology, situate our work within prior efforts, and highlight the advances enabled by our Directed Social Regard (DSR) framework.

\subsection{Terminology}
\label{sec:related-terminology}
We use the following terminology in the remainder of this work. A \textit{text} is a sentence, utterance, or post which may contain one or more \textit{spans}: contiguous sequences of words that may overlap or nest. A span may represent a \textit{topic} (e.g., an event or abstract concept) or a \textit{character} (e.g., a person or group) that is the \textit{target} of a sentiment. \textit{Sentiment} is the opinion or attitude of the author of a text with respect to a target. In most research, sentiment takes values along a polarity dimension (i.e., positive, negative, neutral) and can be categorical or numerical.

Our work introduces a refinement of sentiment we call \textit{regard}. Regard breaks out the concept of sentiment into multiple independent yet complementary dimensions, detailed in \tabref{dimensions_definitions}. These dimensions, rooted in the social sciences, provide a rich characterization of an author's expressed social relationship towards a person, group, or topic. We discuss these dimensions in more detail in \secref{approach}.

\begin{table}[h]
\caption{Definitions of dimensions of regard.}
\label{tab:dimensions_definitions}
\begin{tabular*}{0.75\textwidth}{p{3cm}p{6cm}}
\toprule%
Oppose–Advocate &  Whether the author expresses a desire to hinder or defend a target.  \\
\midrule
Harmful–Helpful &  Whether the author deems the target to be a source of harm or help towards another character or topic.   \\
\midrule
Victimized–Aided &  Whether the author views the target to be a harmed or helped by another character or topic. \\
\botrule
\end{tabular*}
\end{table}


\subsection{Targeted Sentiment Analysis}
\label{sec:related-sentiment}

Targeted Sentiment Analysis seeks to determine the sentiment expressed toward one or more entities mentioned in a sentence. Early work relied on rule-based or feature-driven classifiers, while subsequent work leveraged pretrained transformers (e.g., BERT, RoBERTa) fine-tuned for token classification or span-level tagging \cite{li2019unified,xu2020position}.
Targeted sentiment analysis differs from the simpler task of sentiment classification, which categorically labels an entire sentence as ``negative'' or uses a number to express positive or negative intensity, e.g., $-0.8$ \cite{akhtar2020intense}.

In Aspect-Based Sentiment Analysis (ABSA), models are trained to classify sentiment with respect to specific $\langle$entity, aspect$\rangle$ pairs. For example, in product reviews, ``battery life'' and ``camera'' might be distinct aspects of a phone \cite{pontiki2016semeval}, or that ``leakage'' and ``allergenic'' are aspects of childcare products \cite{yang2018multi}. These methods typically assume the targets and aspects are predefined or extracted via auxiliary tools, and the sentiment polarity is categorical, e.g., with applicable labels \{positive, negative, neutral\}. Some models perform joint extraction of both targets and sentiment labels, but few support multiple overlapping spans or multidimensional evaluation.

Notably, \citeauthor{zheng2022meabsa} \cite{zheng2022meabsa} extend the ABSA paradigm to multi-entity span-based sentiment scoring, introducing a model that assigns real-valued scores along nine sentiment labels to annotated spans in hate speech corpora. Their model uses a frozen transformer encoder and span-wise scoring heads.

Our DSR framework builds on this line of work by (1) automatically extracting multiple spans, (2) scoring each span along multiple interpretable sentiment dimensions supported by social science theories, and (3) allowing for generalization without requiring gold spans or fixed targets. In doing so, our DSR framework extends the ABSA task to a richer and more flexible form of span-targeted sentiment analysis.

\subsection{Stance Detection}
\label{sec:related-stance}

Stance detection is another closely related task, in which the model determines whether the speaker is in favor of, against, or neutral toward an externally-provided target (i.e., proposition, entity, or topic). SemEval-2016 Task 6 \cite{mohammad2016semeval} formalized this problem in Twitter discourse, assuming a fixed target per dataset (e.g., ``Atheism'') and classifying user stance. Subsequent work expanded to multi-target stance detection \cite{xu2019multi} and conversation-level consistency modeling \cite{hazarika2021conversational}, allowing inference of stance across dialogue turns.

Unlike our DSR work, most stance models require the target to be specified in advance and they classify stance as categorical polarity. They typically do not allow for multiple stances in a single utterance and lack support for intensity or multidimensional evaluation. In contrast, DSR identifies targets in the text (via span extraction) and assigns continuous, multidimensional stance scores to each one, capturing subtle differences between condemnation, critique, and opposition. This enables a more faithful representation of nuanced opinion.

Other work on \textit{belief-driven} stance \cite{mather2021general} and concern detection \cite{mather2022stance} produces a predicate-based propositional encoding with NLP semantic role labeling (SRL) \cite{he2017deep}.
For instance, ``that candidate will ruin the economy'' yields a propositional \textit{ruin(candidate, economy)} result with a concern type (e.g., ``economic'') and with numerical scores for belief, sentiment, and/or moral framing \cite{mather2022stance}.
Unlike the above stance-related approaches, this approach can surface concerns and predicates without specifying them \emph{a priori}, and it offers more explainability; however, it also depends on SRL and on a semi-automatic domain-adaptation process to build a moral lexicon for new corpora.
It also differs from DSR by scoring the sentiment of extracted predicate-argument propositions, where DSR labels and scores spans directly.

\subsection{Comparing DSR with other Formulations}

To concretely illustrate the differences between existing approaches and our DSR framework, consider the sentence from \figref{dsr-ui}: ``\textit{How can they do this to the children??}''
We tabulate the various sentiment analysis problem formulations in \tabref{related}, juxtaposing the representative inputs and outputs to illustrate the tradeoffs in expressiveness and constraints.
As shown in \tabref{related}, a standard sentiment classifier would assign a single global label (e.g., ``negative'') to the entire sentence, without identifying to whom the sentiment is directed, and the intensity scoring likewise generates a blanket document-level output.
These sentence-level outputs are not expressive enough to capture the sentence's targets, e.g., the positive sentiment targeted at ``children'' and the negative sentiment targeted at ``they'' and at the under-specified ``do this.''

Similarly, a sentiment intensity scoring approach (\tabref{related}, second row) could produce a more nuanced real-valued number, but this incurs the same singular-output problem as sentiment classification: a single number does not adequately communicate the presence of both advocacy for children and opposition to the harmful act.
Emotion scoring has successfully scored multiple, conflicting dimensions of emotion in a single message \cite{guo2022emotion}, but likewise does not attribute these emotions to various targets or topics in the message.

\begin{table}[h]
\caption{Comparing representative inputs and outputs of related sentiment and stance analysis techniques with the DSR approach presented in this paper.
}
\label{tab:related}
\begin{tabular*}{\textwidth}{p{3cm}p{5cm}p{4cm}}
\toprule%
\textbf{Formulation} & \textbf{Representative Input(s)} & \textbf{Representative Output(s)} \\ \toprule
Sentiment Classification &
\textbf{Text}: ``How could they do this to the children??'' &
\textit{negative} \\ \midrule
Sentiment Intensity Scoring \cite{akhtar2020intense} &
\textbf{Text}: ``How could they do this to the children??'' &
$-0.8$ \\ \midrule
\multirow{7}{*}{Stance Detection \cite{mohammad2016semeval}} &
\textbf{Text}: ``How could they do this to the children??''; \textbf{Target}: ``Children'' &
\textit{positive} \\ \cmidrule(lr){2-3}
 &
\textbf{Text}: ``How could they do this to the children??''; \textbf{Target}: ``They'' &
\textit{negative} \\ \cmidrule(lr){2-3}
 &
\textbf{Text}: ``How could they do this to the children??''; \textbf{Target}: ``Restaurants'' &
\textit{neutral} \\ \midrule
\multirow{4}{3cm}{Multi-Entity ABSA \cite{yang2018multi}} &
\textbf{Text}: ``How could they do this to the children??'';
\textbf{Entities}: \{``they'', ``children''\};
\textbf{Aspects}: \{``harmful'', ``harmed''\}
&
\textbf{children}: \par
harmful=$neu$, harmed=$pos$;\par
\textbf{they}:\par
harmful=$pos$, harmed=$neu$;
\\ \midrule
\multirow{6}{3cm}{Directed Social Regard (Present Approach)} &
\multirow{6}{5cm}{\textbf{Text}: ``How could they do this to the children??''} &
\textbf{they} (\textit{Character}): \par
oppo-advo=$-.57$ (Opposed)\par
victm-aided=$-.01$ (Neutral)\par
harmf-helpf=$-.51$ (Harmful);\par
\textbf{do this} (\textit{Topic}): ... \par
\textbf{children} (\textit{Character}): ...
\\ \botrule
\end{tabular*}
\end{table}

Traditional stance detection methods (\tabref{related}, rows 3-5) can detect targeted sentiment, but they require the target specified \emph{a priori} by a user or by some antecedent target-detection operation.
In this formulation, if we know the full set of relevant sentiment targets at the onset of stance detection (e.g., we are only searching for sentiment targeted at ``children'' or at pronouns such as ``they''), then this approach will allow us to run a sequence of sentiment classifications for each target. However, we do not often know the sentiment targets \emph{a priori}, especially when we analyze novel, broad corpora.
Moreover, suppose that ``children'' are mentioned more than once in the text, once with advocacy and once with opposition.
Stance approaches would not be able to capture the duality of this sentiment, since it only predicts a single sentiment per target.

An aspect-based sentiment model (\tabref{related}, row six) might require pre-specified targets and pre-specified aspects, assuming the system is capable of extracting both entities. Targeted sentiment models extend this by allowing per-target sentiment labeling, but typically assume the targets are known ahead of time and lack the ability to express multidimensional or nuanced attitudes.
This approach is more expressive than stance detection, but it faces similar limitations of requiring the targets to be specified up-front.

As shown in the bottom row of \tabref{related}, our DSR approach is given only the raw text and it automatically identifies salient spans such as ``they,'' ``do this,'' and ``children,'' and then assigns each a category (Character or Topic) and real-valued scores for interpretable sentiment dimensions of Opposed–Advocated, Victimized–Aided, and Harmful–Helpful.
This enables DSR to offer a more granular, multidimensional account of sentiment, grounded in learned span representations and without requiring fixed targets or predefined aspect labels.
As we discuss later, these resulting DSR attributes have composite meaning, e.g., expressing that ``they'' are both Opposed and Harmful has a macro-level effect of opposing harms to others. However, as we will see in our applications on hate speech corpora in \secref{application}, we can use DSR to quickly identify advocacy for harms and violence.

%% file: approach.tex
\section{Approach}
\label{sec:approach}

This section starts with an explanation of the social science theories backing DSR and how each dimension was determined. It then describes the construction of the DSR dataset in \secref{data} from data collection through annotation. Finally, \secref{model} describes our NLP models for span recognition and regard scoring.

\subsection{Supporting social science theories}
\label{sec:socialscience}

DSR has theoretical foundations in social science theories that have been applied across cultures and social contexts.
We outline two salient theories of regard in language and then connect them to the three dimensions of regard utilized in this work.

Brown and Levinson's \cite{brown1987politeness} cross-cultural linguistic theory of politeness and etiquette \cite{brown1987politeness} holds that language is a means to negotiate social relationships.
According to their theory, language is a mitigation strategy to maintain a positive self-image (or ``face'') in public communication with others, and language can direct complex sentiment at others that may cause them to gain or lose face.
According to this theory, language can affect face whenever it refers to a person, their ideas, aspects of their identity, or their property and possessions.
These considerations impact our decision to represent ``character'' as a type of span in our DSR framework to cover all of these elements. For possessions (e.g., of ideas and physical property) we also capture possessive pronouns such as ``my'', ``your'', and ``hers''.


Albert Bandura's \cite{bandura1996mechanisms,bandura2000exercise} psychosocial theory of moral disengagement focuses on the cognitive mechanisms and linguistic indicators related to our moral justifications for doing harm or good, assigning positive or negative responsibility, and including or excluding others in our circle of humanity.
Bandura's theory includes a taxonomy of linguistic cues that indicate moral justification or disengagement, such as ascribing blame, expressing victimization, justifying harms, in-grouping, and out-grouping, all of which directly motivate our DSR dimensions (described below).
Under this theory, the power comes from the collective behavior.
These individual indicators become amplified at the group level through what Bandura termed collective efficacy—shared beliefs in a group's collective power to produce desired results.
When groups display high collective efficacy, they become more confident and more likely to advocate strongly for their positions, often employing moral disengagement strategies collectively to justify their actions.
This suggests that DSR is especially useful for summarizing group- or community-level indicators, which we exemplify in our corpus analysis experiments.

We next review the three dimensions of DSR annotated and modeled in this work, including their social science motivations an theoretical support.
Although these concepts derive from social theory, they operate as fundamentally relational.
Attitudes toward others---whether positive or negative---are targeted orientations toward specific groups rather than abstract sentiments, thereby comprising \emph{directed} social regard.

\subsubsection{Oppose--Advocate}

Per Brown and Levinson \cite{brown1987politeness}, language can recognize and respect other's social standing and dignity (e.g., advocate for them and their ideas and possessions) or threaten it.
Consequently, capturing the positive (i.e., \emph{advocacy}) and negative (i.e., \emph{opposition}) valence and the target of the regard helps model this core component of Brown and Levinson's theory: directed regard of Opposition-Advocacy helps predict an author's attempt to save face or threaten the face of others.

According to Bandura \cite{bandura1996mechanisms,bandura2000exercise}, we do not apply our moral standards uniformly.
When we advocate for our in-group, we may selectively activate moral standards that favor our own group or ideas and disengage from our moral standards when considering the concerns of out-groups.
Group dynamics are often manipulated by justifying why an advocated group deserves a positive position and an opposed group deserves its inferior position.
This theory mentions \emph{dehumanization} as an extreme opposition strategy, by excluding people from our circle of humanity (and thus from moral consideration).
Dehumanizing language may include derogatory labels like ``cockroaches'' or ``vermin'' and effectively removes the moral censure of hostile behavior toward dehumanized groups.
In our annotation guidelines, we exemplify dehumanization as extreme opposition.



Other theoretical perspectives examine how advocacy and opposition in social systems drive group identity and group behavior.  System Justification Theory states that people tend to defend and justify the system they are part of, even when those systems disadvantage them. The concept extends to group advocacy, when people rationalize their group's actions as legitimate, even when such rationalization lacks objective justification \cite{jost2012system}. Both advocacy and opposition serve a social function. In political discourse, it often shapes our political identity: we define who we are by defining who you oppose \cite{simmel1904sociology}.
Since identity is shaped by values and culture, these become focal points of inter-group opposition: what one group sees as advocacy another sees as moral opposition \cite{haidt2012righteous}.
In sum, advocacy and opposition are best understood as complementary processes grounded in social theory, where advocacy of persons or actions should be understood as positive directed social regard and opposition as negative directed social regard, and mechanisms of moral disengagement enable both simultaneously.

\subsubsection{Harmful--Helpful}

Blaming others for harm is one way to condemn them as an aggressor or as an immoral agent, and it can help morally justify subsequent adverse acts against them in return \cite{bandura1996mechanisms,bandura2000exercise}.
This attribution of harmfulness occurs frequently in geopolitical messaging, where the harms of one group (against another) are expressed to help justify interventions against them.

According to Bandura \cite{bandura2002selective}, blame attribution can likewise displace or diffuse responsibility onto others, allowing the speaker to avoid or diminish blame toward themselves.
This includes the strategy of blaming-the-victim, where a group or individual may be both opposed and victimized.
This may justify mistreatment of another by saying that the person ``deserved it'' or was ``just asking for it.''
This composite, multi-dimensional regard is a case for observing multiple dimensions in conjunction, which we exemplify in our analyses.


Weiner's attribution theory shows a different set of cognitive mechanisms underlying both harming and helping \cite{weiner1995judgments}. When individuals encounter others in need, they engage in causal attributions about the reasons for that need. Attributions to controllable causes (personal responsibility, poor choices) tend to evoke blame, anger, and reduced helping, while attributions to uncontrollable causes (circumstances, bad luck) evoke sympathy and increased helping behavior. This attribution process determines whether observers extend or withdraw their advocacy. 

Jost's system justification theory shows how blaming serves ideological functions \cite{jost2012system}. Blaming individuals for their circumstances helps justify existing social systems by suggesting that inequalities reflect personal failings rather than structural problems. This allows actors to avoid the cognitive dissonance that would arise from recognizing systemic injustice while simultaneously benefiting from or participating in those systems.

Batson's work demonstrates that genuine altruistic helping is often motivated by empathetic concern for others' welfare, requiring the helper to maintain rather than suspend recognition of their humanity \cite{batson2011altruism}, suggesting that when individuals genuinely empathize with those in need, they are motivated to help regardless of personal cost or benefit. And finally, Gilligan's ethics of care framework provides another lens for understanding helping behaviors as positive social regard \cite{gilligan1982different}. Where traditional moral frameworks focus on rights, the care perspective emphasizes relationships, responsibility, and attention to particular others' needs. 
Consequently, expressing a group as helpful is an avenue to humanizing them.


\subsubsection{Victimized--Aided}

Victimization may be expressed in the presence of an explicit aggressor (e.g., where one group harms another) or absent an aggressor (e.g., where life is difficult or unmanageable for a group or individual).
Conversely, regarding a character as aided is describing them as benefiting in some fashion, with or without the mention of a helpful agent.
Both of these can apply to the speaker themselves, e.g., if the speaker regards themselves as victims of an unfair system or as blessed by a turn of events.

Expressions of harm and care for individuals is also a central component of Moral Foundations Theory (MFT) \cite{haidt2012righteous,graham2013moral}.
MFT acknowledges that there are a multitude of ways to trigger feelings of compassion for victims, and that this compassion is often mixed with anger toward those who cause harm \cite{graham2013moral}. 
By modeling both the victimized regard for a group and the harmful regard (above) for an aggressor, DSR is able to capture this pairwise relationship in a single context, which we demonstrate in our extraction of pairwise themes in \secref{app-pairwise}.

\subsubsection{Importance of layering multiple dimensions of DSR}

As described above, the three dimensions of social regard extracted in this work are interrelated in social science theories.
When a character or topic is regarded as both harmful and opposed, it may be an intent to diminish the moral standing of that individual \cite{brown1987politeness} or justify subsequent action against them \cite{bandura1996mechanisms}.
Conversely, if a character is described as both harmful and advocated, the harms may be justified by the speaker \cite{bandura1996mechanisms}, which could indicate an escalation of harmful actions.
Similarly, if a character is described as victimized and opposed or harmful, this may be a case of shifting blame to the victim or diminishing the suffering of the victim \cite{bandura1996mechanisms}.

When combined, these indicators help distinguish whether the harms acknowledged in a message are justified (strongly advocated), atrocities (strongly opposed), or being dryly reported (neutral), and they help distinguish whether a harmed or suffering character is being deeply empathized with (strongly advocated) or being diminished or dehumanized in their victimization (opposed).
We demonstrate the extraction and comparison of multiple DSR indicators in \secref{app-composite}.
These combined dimensions of social regard provide a descriptive, directed account of sentiment and micro-narratives in language.

\input{data}

\input{model}

%% file: data.tex
\subsection{Constructing labeled Directed Social Regard datasets}
\label{sec:data}

To train and evaluate our DSR approach, we collected publicly accessible information (PAI) from multiple social media platforms and then worked with trained temp workers to annotate it with DSR categories over a period of eight weeks.
Each text in our dataset has an average of 7.5 textual spans, and each of these spans are scored for 2-3 dimensions of DSR intensity, yielding an average of 14.5 overlapping scores from at least two annotators on each text.
We identified large datasets for collection (\secref{collection}) but had to randomly down-sample these aggressively due to the amount of human annotator effort required for each text.
Our final DSR dataset contains a total of 1,838 texts from eight diverse sources, comprising 13,893 character and topic spans that have been scored with a total of 26,779 averaged, high-agreement intensity scores across annotators.
We describe the data collection, data annotation, and data curation processes next.

\subsubsection{Data Collection}
\label{sec:collection}
We collected PAI data from existing datasets and online sources, with the goal of accruing cross-domain expressions of advocacy, opposition, harms, care, [dis]respect, and neutral sentiment.
PAI data were sampled from the sources listed in \tabref{dsrdata}, spanning multiple forums and social media platforms, years, domains of discussion, and countries of origin.

\begin{table}[h]
\begin{minipage}{\linewidth}
\caption{Publicly-accessible information (PAI) sources sampled to generate the DSR dataset for training and testing.}
\label{tab:dsrdata}
\begin{tabular*}{\textwidth}{lp{10cm}}
\toprule%
\textbf{Rand. Sample} & \textbf{Dataset Description} 
\\ \midrule
489 Posts &
Pushshift Reddit dataset \cite{baumgartner2020pushshift}, including a selection of subreddits that contain attributes of interest for social regard. 
\\ \midrule
450 Comments &
New York Times Annotated Corpus\footnote{https://catalog.ldc.upenn.edu/docs/LDC2008T19/new\_york\_times\_annotated\_corpus.pdf} comprising comments and news articles. 
\\ \midrule
379 Posts &
Social media dataset from 2021 annotated for harms, empathy, dehumanization, and other indicators of moral [dis]engagement \cite{friedman2021toward}. 
\\ \midrule
335 Tweets &
Kaggle dataset of English tweets about the 2017 French presidential election.\footnote{https://www.kaggle.com/datasets/jeanmidev/french-presidential-election} 
\\ \midrule
52 Tweets &
Moral Foundations Twitter Corpus (MFTC) \cite{hoover2020moral}\footnote{https://osf.io/k5n7y/} comprising tweets from seven distinct domains of discourse---such as natural disasters, politics, and contemporary social issues---that have been hand-annotated by at least three trained annotators for 10 categories of moral sentiment.
\\ \midrule
52 Posts &
The Gab Corpus \cite{kennedy2018gab}\footnote{https://osf.io/edua3/} comprising posts from the Gab social network, annotated for rhetoric by a minimum of 3 annotators.
Its coding typology includes hierarchical labels indicating dehumanizing and violent speech as well as indicators of targeted groups and rhetorical framing.
\\ \midrule
44 Tweets &
The Kaggle ``How ISIS Uses Twitter'' dataset\footnote{https://www.kaggle.com/datasets/fifthtribe/how-isis-uses-twitter} that has been utilized for research on detection of radical social media detection \cite{ul2021understanding}, comprising tweets from 100+ pro-ISIS accounts since the November 2015 Paris Attacks.
\\ \midrule
37 Posts &
Online forums provided by Ribeiro et al.'s (\citeyear{ribeiro2021evolution}) analysis of the ``Manosphere,'' frequently including expressions of gender inequality, self-victimization, and discontent. \\
\botrule
\end{tabular*}
\end{minipage}
\end{table}

As mentioned above, our annotation process (\secref{annotation}) involved substantial human annotator attention to every phrase of interest within each example text, so the final dataset is a random sample of the of the diverse sources in \tabref{dsrdata}, totaling 1,838 texts from eight PAI sources.

\subsubsection{Data Preparation}

Three NLP experts used a web-based annotation interface to annotate the characters and topics in each message, comprising the dataset for the DSR token classification model (see \secref{eval-ner}).
The DSR token classification model was used in a human-machine model-in-the-loop strategy, running the token-classification model and then extending or revising the resulting spans with human expert judgment.
From the 1,838 texts collected (\secref{collection}), this preparation phase produced a total of 13,893 spans to annotate (9,403 character spans and 4,490 topic spans).

\subsubsection{Data Annotation}
\label{sec:annotation}
Given the potentially toxic content and the potential mental health risks of reading these texts, we obtained separate IRB and HRPO approvals to conduct the annotation study.
All participants were required to sign a consent form before beginning the work.
Researchers collaborated with a staffing agency to recruit four temporary part-time workers to perform the annotations.
Workers were selected based on their interview responses and their ability to clearly articulate their thought process and rationale when annotating examples.
Annotators were compensated at \$15 per hour and hired for an 8-week period, with a workload of roughly 20 hours per week.

\begin{figure}[h]
\centering
\includegraphics[width=.8\textwidth]{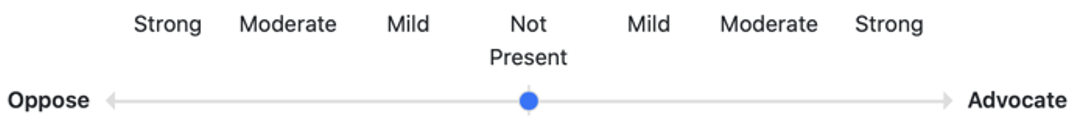}
\caption{Single-dimension slider used by annotators to indicate a rating from strong Oppose (left) to strong Advocate (right) for every character or topic span, in context.  Similar sliders were used for Harmful-Helpful and Victimized-Aided dimensions.}
\label{fig:slider}
\end{figure}

We utilized an annotator-in-the-loop workflow \cite{schmer2024annotator}, conducting group discussion sessions with annotators and research staff.
In these sessions, annotators and researchers collaboratively refined DSR concept definitions, worked through exemplary edge cases, and answered questions.
Outside of sessions, annotators accessed a web-based annotation interface that highlights each span (i.e., character or topic) within the context of the text and prompts the annotator for an Opposed-Advocated score (for character spans and topic spans) and a Harmful-Helpful score and Victimized-Aided score (for character spans only).
Scores were entered using a slider as shown in \figref{slider}, where `Strong'' to the left and right yields an intensity score of -1.0 and 1.0, respectively, and the default ``Not Present'' yields an intensity score of 0.0.

\subsubsection{Dataset Properties and Annotator Agreement}
\label{sec:agreement}

We conducted a manual filtering phase to remove individual intensity scores that did not accord with the annotation instructions.
We then sorted ratings for each span by their standard deviation over annotators' scores, and detected and reviewed outliers that may have corresponded to mistakes (e.g., inflections over the x-axis), and escalated these for possible filtering.
After this mixed semi-automatic rectification, we arrived at the span and attribute counts reported above.
\tabref{agreement} reports the Krippendorff's $\alpha$ for each attribute after filtering, showing that each of the regard dimensions are within an acceptable range.

\begin{table}[h]
\caption{Krippendorff's $\alpha$ for DSR attributes in the labeled training/testing data.}
\label{tab:agreement}
\begin{tabular*}{.5\textwidth}{lcc}
\toprule%
\textbf{Attribute} & \textbf{\# Scores} & \textbf{$a$}
\\ \midrule
Oppose–Advocate & 10,101 & 0.91 \\
Harmful–Helpful & 8,327 & 0.71 \\
Victimized–Aided & 8,351 & 0.73
\\ \midrule
Total \& Micro-Avg. & 26,779 & 0.87 
\\ \botrule
\end{tabular*}
\end{table}

\begin{figure}[h]
\centering
\begin{minipage}[t]{0.38\textwidth}
    \includegraphics[width=\textwidth]{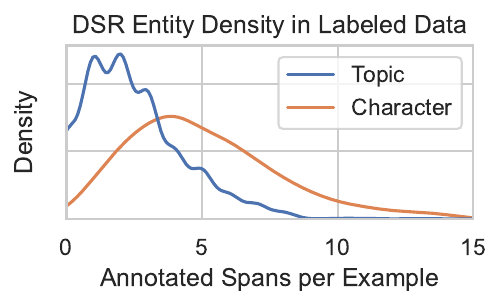}
    \caption{Spans per example: \textit{topics} (non-human subjects) and \textit{characters} (human individuals or groups).}
    \label{fig:ent-density}
\end{minipage}\hfill
\begin{minipage}[t]{0.58\textwidth}
    \includegraphics[width=\textwidth]{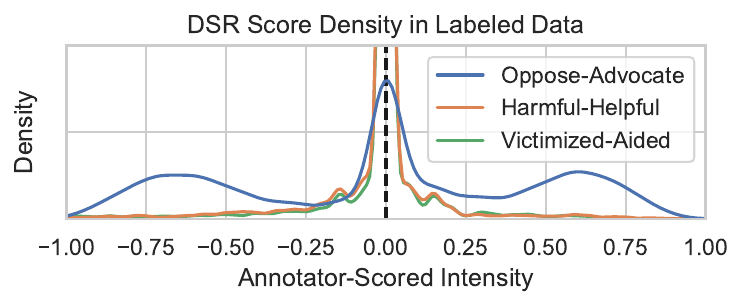}
    \caption{Score density for all three DSR dimensions, from negative (Oppose, Harmful, Victimized) to positive (Advocate, Helpful, Aided).}
    \label{fig:attr-density}
\end{minipage}
\end{figure}

We plot the entity density and score density of the resulting dataset in \figref{ent-density} and \figref{attr-density}, respectively.
\figref{ent-density} shows that character spans occur with higher density in texts than non-human topics of regard.
\figref{attr-density} shows a truncated view of the density of intensity scores, since many spans were scored as having near-neutral Harmful–Helpful and Victimized–Aided dimensions.
The negative (Opposed, Harmful, Victimized) and positive (Advocated, Helpful, Aided) axes of \figref{attr-density} are generally balanced, with equivalent representation of both poles of regard for each dimension, and many neutrals.

%% file: model.tex
\subsection{NLP/ML Models}
\label{sec:model}

The DSR NLP capability is split into two separate models: a token classification model (\secref{token-classifier}) followed by a separate span-level intensity-scoring model (\secref{span-scorer}).
We describe these models in the following sections, and we present our validation results for both of these models in \secref{results}.

\subsubsection{Token Classification}
\label{sec:token-classifier}

Token classification is a well-studied problem in NLP, and off-the-shelf architectures\footnote{https://huggingface.co/docs/transformers/en/tasks/token\_classification} provide state-of-the-art performance after fine-tuning on domain-specific datasets.
Our DSR token classification approach therefore uses the open-source HuggingFace Transformers library's TokenClassification interface.


The input to this NLP model is unstructured text and the output is a character and topic labeling of spans over the text as shown in \figref{dsr-ui}, including single-word spans (e.g., the characters \textq{they} and \textq{children}) and multi-word spans (e.g., the topic \textq{do this}). The output of the token classification model is passed as input to the regard scoring model, described next.

\subsubsection{Regard Scoring}
\label{sec:span-scorer}

\begin{figure}[h]
\centering
\includegraphics[width=\textwidth]{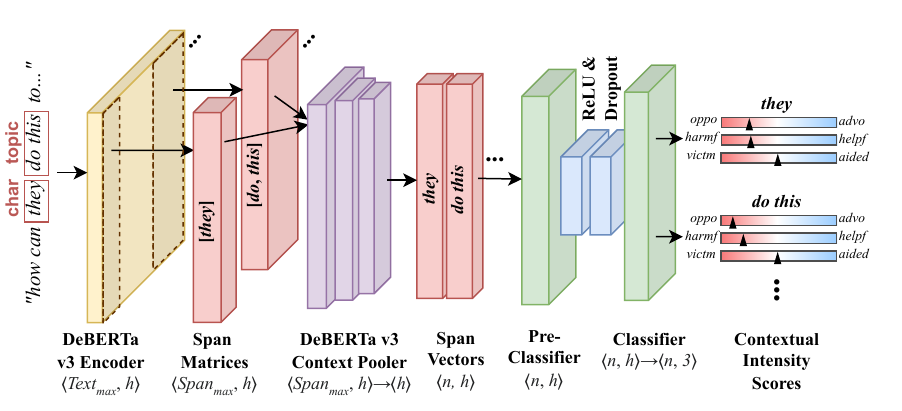}
\vspace{-.2in}
\caption{Model diagram of the best-performing contextual span-scoring model.  Text with target spans (left) is processed by a transformer-based encoder, span-specific matrices are extracted and pooled contextually, culminating in a three-dimensional linear layer that outputs a per-span [-1, 1] logit for each DSR dimension.}
\label{fig:model}
\end{figure}

The diagram in \figref{model} shows the architecture of the regard scoring model, implemented in Python using the PyTorch and HuggingFace Transformers libraries.
The dimensions of the tensor-based components in \figref{model} are specified with variables for the hidden layer size $h=1024$, the maximum text length in words $Text_{max}=512$, the maximum span count per text $Span_{max}=200$, and the number of spans in a text $n$.

The model encodes each text exactly once (\figref{model}, left), aggregates context-sensitive matrices for each span (i.e., Character or Topic) in the text, and uses contextual pooling to normalize each single- or multi-word phrase into a uniform tensor of length $h$.
It batches span vectors through linear and activation layers, culminating in a classifier that produces a logit for each of the three DSR dimensions (Oppose–Advocate, Harmful–Helpful, and Victimized–Aided) for each span labeled in the input text.

Since this is a regression-based model, we use mean-squared error (MSE) as the loss function, where $N$ is the number of data points, $y_i$ is the desired score, and $\hat{y}_i$ is the corresponding model-predicted score:

\begin{equation}
MSE = \frac{1}{N} \sum_{i=1}^{N}(y_i - \hat{y}_i)^2
\end{equation}


As described in our data annotation process (\secref{annotation}), we did not collect Harmful–Helpful or Victimized–Aided scores for non-human topic spans, since our social science theories primarily deal with humans.
This means that while the model \textit{does} generate Harmful–Helpful and Victimized–Aided scores for topics, these scores are \textit{not} factored into the loss function during training, since we do not have human labels for these dimensions.

%% file: results.tex
\section{Experiments}
\label{sec:results}

We first present the performance of our span recognition model in \secref{eval-ner}.
We then evaluate our regard scoring model in \secref{eval-absa}.
Both of our validation experiments split the respective datasets into a 90\% train split and a 10\% held-out test split.


\subsection{Span Recognition}
\label{sec:eval-ner}

Consistent with previous work in this area, we use per-label and micro-averaged metrics of precision ($p$), recall ($r$), and F1 score.
We treat the base model as an independent variable to observe its effect.
For all models we train for 10 epochs with a batch size of 8 documents, 1 gradient accumulation step, and we vary the learning rate per the model type based on best practices (5e-5 for BERT variants, 6e-6 for RoBERTa and DeBERTa).
In total, there were 914 character spans to recognize and 443 topic spans to recognize in the test set.

\begin{table}[h]
\caption{Results of using different base models for training and evaluating token classification of character spans and topical spans in a 10\% hold-out test split. Metrics include precision ($p$), recall ($r$), and F1 score, where higher is better.}
\label{tab:ner}
\begin{tabular*}{\textwidth}{@{\extracolsep\fill}lcccccccccccc}
\toprule%
& \multicolumn{9}{@{}c@{}}{Span-Level (SemEval Strict)} & \multicolumn{3}{@{}c@{}}{Token-Level} \\
\cmidrule(lr){2-10} \cmidrule(lr){11-13}%
& \multicolumn{3}{@{}c@{}}{Character} & \multicolumn{3}{@{}c@{}}{Topic} & \multicolumn{3}{@{}c@{}}{Micro Avg.} & \multicolumn{3}{@{}c@{}}{Micro Avg.} \\
\cmidrule(lr){2-4} \cmidrule(lr){5-7} \cmidrule(lr){8-10} \cmidrule(lr){11-13}%
Base Model         & $p$ & $r$ &  F1 & $p$ & $r$ &  F1 & $p$ & $r$ & F1  & $p$ & $r$ &  F1 \\ \midrule
distilbert         & .94 & .95 & .95 & .69 & .70 & .69 & .86 & .87 & .86 & .90 & .88 & .89 \\
bert-base-uncased  & .95 & .94 & .95 & .69 & .70 & .70 & .86 & .87 & .86 & .89 & .88 & .89 \\
multilingual-bert  & .94 & .94 & .94 & .72 & .66 & .69 & .85 & .86 & .86 &.91& .87 & .88 \\
xlm-roberta-lg     & .94 & .94 & .94 & .71 & .69 & .70 & .87 & .86 & .86 &.91&.89& .90\\ \midrule
deberta-v3         & .95 &\tb{.96}&\tb{.96}& .74 & .72 & .73 & .88 & .88 & .88 &.91 &\tb{.90} & .90\\
deberta-v3-mnli    & .95 & .95 & .95 &.74&.72&.73&.88&.88&.88&.91&.89& .90\\
deberta-v3-lg      &\tb{.96}&\tb{.96}&\tb{.96}&.73 &.70& .71 &.88& .87 &.88&.91&.89& .90 \\
deberta-v3-lg-mnli &.95&\tb{.96}&\tb{.96}&\tb{.75}&\tb{.74}&\tb{.74}&\tb{.89}&\tb{.89}&\tb{.89}&\tb{.93}&\tb{.90}&\tb{.92}\\
\botrule
\end{tabular*}
\end{table}

We report results in \tabref{ner}, splitting out the results for character spans, topic spans, a micro-average of span results, and a micro-average of token results.
The first three of these require \emph{exact span match} to be correct, equivalent to the ``Strict'' setting of SemEval-14 \cite{pradhan2014semeval}.
Under this setting, if a the model predicts a character span \textq{all E.U. countries} when the intended span is the shorter sub-span \textq{E.U. countries,} then no partial credit is given.
The rightmost column in \tabref{ner} provides partial credit for matching, similar to the ``Relaxed'' SemEval setting \cite{pradhan2014semeval}.

As shown in \tabref{ner}, the DeBERTa-v3 variants \cite{he2020deberta,he2021debertav3} outperform other base models by a small margin, and the open-source large DeBERTa V3 pre-trained on various NLI datasets\footnote{https://huggingface.co/MoritzLaurer/DeBERTa-v3-large-mnli-fever-anli-ling-wanli} was the highest performing base model overall.
As expected, the token-level metrics yield higher performance measures than the strict span-level metrics due to the partial correctness for overlapping span predictions.

The difference in performance recognizing character spans (F1 $ \approx 0.96$) vs. Topic spans (F1 $ \approx 0.74$) across base models under the strict setting illustrates the relative difficulty of these tasks.
Contributing factors for this difference include (1) the relative ease of identifying some characters, e.g., due to recurring human pronouns and proper nouns for people and organizations, (2) the relative difficulty of strictly predicting the bounds of an event or entity topic, e.g., \textq{bilateral peace treaty} vs. \textq{peace treaty,} and (3) the relative difficulty of predicting whether a topical phrase is of enough contextual importance to infer it as a Topic.
All else being equal, we value high Recall (i.e., completeness) in Topic detection over high Precision, since high recall will provide more spans to the regard scoring model.

Altogether, the performance of the best-performing models, with F1 $= 0.89$ under strict correctness and F1 $= 0.92$ under partial correctness, validate its capability for our social science use-cases.

\subsection{Regard Scoring}
\label{sec:eval-absa}

We conducted an isolated evaluation of the regard scoring model to assess its capability independently of span recognition.
We held out 10\% of the human-scored texts along the three DSR dimensions of interest and then used the model to predict those scores, computing RMSE to measure error and $R^2$ to measure correlation.
The test set had 1043 ratings of Opposed--Advocated, 870 ratings of Victimized--Aided, and 840 ratings of Harmful--Helpful.
The difference in ratings per dimension is due to our filtering out of low-agreement spans as described above. Similar to the span-recognition experiment, we vary the base model to establish a range of performance and identify a best-performing variant.
In each condition, we train the model for 20 iterations using the RMSE loss function.

\begin{table}[h]
\caption{Results of regard scoring across models for RMSE error metric (lower is better) and $R^2$ correlation metric (higher is better).}
\label{tab:absa}
\begin{tabular*}{\textwidth}{@{\extracolsep\fill}lcccccc}
\toprule%
& \multicolumn{2}{@{}c@{}}{Opposed--Advocated} & \multicolumn{2}{@{}c@{}}{Victimized--Aided} & \multicolumn{2}{@{}c@{}}{Harmful--Helpful} \\\cmidrule(lr){2-3}\cmidrule(lr){4-5}\cmidrule(lr){6-7}%
Base Model                  & RMSE   & $R^2$  & RMSE   & $R^2$  & RMSE   & $R^2$  \\
\midrule
bert-base-uncased           & .29    & .68    & .12    & .55    & .15    & .37    \\
modernbert-lg               & .25    & .77    & .11    & .58    & .13    & .46    \\
deberta-v3-sm               & .26    & .75    & .11    & .58    & .14    & .42    \\ \midrule
deberta-v3-base             & .23    & .80    & .11    & .57    & .13    & .46    \\
deberta-v3-base, debiased   & .22    & .81    & .11    & .62    & .13    & .53    \\ \midrule
deberta-v3-lg               &\tb{.20}& .84    &\tb{.10}& .66    &\tb{.12}& .45    \\
deberta-v3-lg, debiased     &\tb{.20}&\tb{.85}&\tb{.10}&\tb{.67}&\tb{.12}&\tb{.57}\\ \midrule
Llama 3.2 3B Instruct (LoRA)& .26    & .75 & - & - & - & - \\
GPT-4o (few-shot)           & .40    & .39 & - & - & - & - \\
\botrule
\end{tabular*}
\end{table}

We performed additional experiments with the DeBERTa-v3 models using a norm-based data augmentation method \cite{friedman2024debiasing}, denoted as \textq{debiased} in \tabref{absa}.
This augments the training data with additional, template-based examples that vary the characters (e.g., \textq{Russia,} \textq{China,} \textq{Nigeria,} \textq{World Health Organization,} etc.) and the topics (e.g., \textq{free speech,} \textq{immigration restrictions,} etc.) while keeping the rest of the message context and scores constant.
This is designed to help the model attend to context instead of the span-content so that it doesn’t sub-optimally associate a context-free negative (or positive) regard with a concept that is frequently opposed (or advocated) in the training data, but may be advocated (or opposed) in other corpora or domains.
As shown in \tabref{absa}, the debiasing operation increased correlative measure ($R^2$) in all dimensions and it decreased or maintained the error measure (RMSE) in all dimensions.

\subsubsection{Comparison to LLM performance}

For comparison with much larger open-source and commercial large language models (LLMs), we fine-tuned a Llama 3.2 Instruct base LLM\footnote{https://huggingface.co/meta-llama/Llama-3.2-3B-Instruct} for this task, and we conducted a few-shot trial with OpenAI's GPT-4o model.
For these LLM conditions, we used only the Opposed-Advocated scores for simplicity.
To adapt the regard scoring task to a text-to-text LLM setting, we created a prompt instructing the LLM score various spans from -100 (highly opposed) to 100 (highly advocated), with 0 being neutral, and we extracted the numbers from the LLM's output.
For the Llama 3.2 condition, we used the same train/test split and fine-tuned Llama 3.2 using low-rank adaptation (LoRA) \cite{hu2021lora}.
\tabref{absa} shows that the LoRA-trained Llama LLM performs on par with the smaller DeBERTa-based models, but it is substantially outperformed by the larger DeBERTa variants.
The Llama 3.2 base LLM has 3 billion parameters (over 600\% more parameters than the DeBERTa-v3-large models).

The GPT-4o condition was conducted in a few-shot setup by providing 10 worked examples with each request and then decoding the results.
As shown in \tabref{absa}, this is the lowest-performing condition, most likely because the model received orders of magnitude fewer examples to learn the nuances of the task.

Both the fine-tuned Llama 3.2 and the few-shot GPT-4o models underperformed smaller purpose-built transformer-based models.
A likely reason is that the DSR training for the transformer-based models (DeBERTa, etc.) all used RMSE loss functions that guided learning based on the numerical error of their predictions; however, the text-to-text Llama 3.2 and GPT-4o LLMs had to encode and decode numbers within text rather than receive the numerical data directly to and from their neural layers.
Both LLM conditions show that LLMs can be fine-tuned or prompted to perform simple variants of this task, but smaller purpose-built models outperform them at present.

\begin{figure}[h]
\centering
\includegraphics[width=\textwidth]{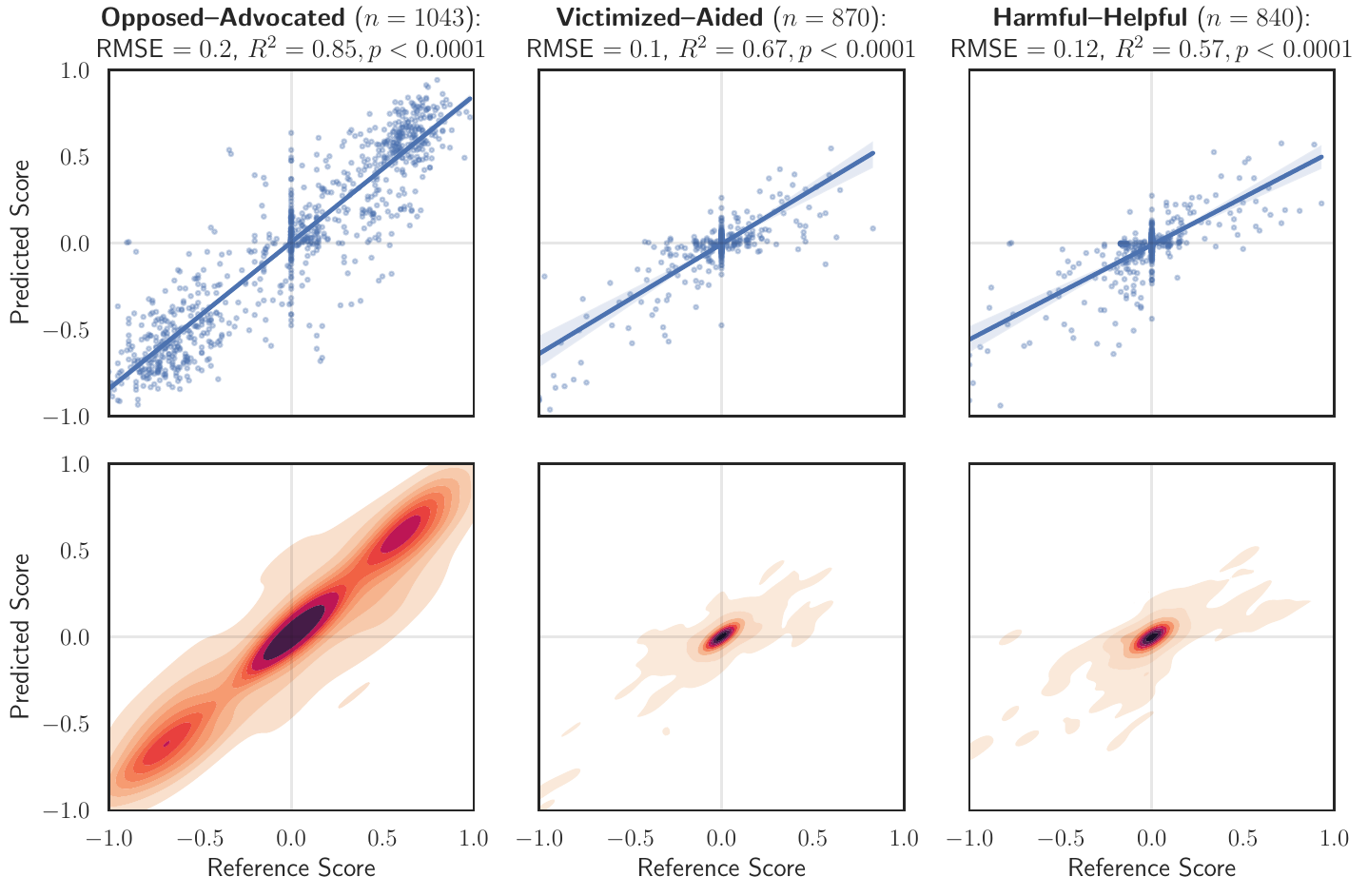}
\caption{Scatter plots (top) and density plots (bottom) for all three dimensions of regard, plotting human reference scores ($x$-axes)against predictions of the best-performing DeBERTa-v3-large debiased model ($y$-axes).}
\label{fig:absa-plot}
\end{figure}

\subsubsection{Density and distribution of regard scores}

Scatter plots and density plots for all three dimensions are shown in \figref{absa-plot}, plotting annotators' reference scores against the predictions of the best-performing (DeBERTa-v3-large, debiased) variant of the DSR intensity-scorer.
The density plots show high density around the origin, indicating neutral regard in the reference data (i.e., $x \approx 0$) and accordingly in the predictions (i.e., $y \approx 0$).
This shows that the test split contains a high incidence of neutral regard in addition to some extremes, and the model can suitably capture the variation in the data, as evidenced by the favorable $R^2$ scores.

\subsection{Applying DSR on existing datasets}
\label{sec:application}

To further demonstrate and validate the DSR approach, we apply it to data from pre-existing PAI datasets that have been curated by other researchers and annotated by trained raters for other sociolinguistic attributes.
This allows us to characterize how DSR results relate to other sociolinguistic attributes and demonstrate how DSR helps surface distinctive themes within various datasets.
All pre-existing sources and their dataset sizes are listed in \tabref{appdata}.
This NLP research effort did not involve any additional collection of PAI data from online sources.

\begin{table}[h]
\begin{minipage}{\linewidth}
\caption{Pre-existing publicly-accessible information (PAI) datasets comprising the DSR application studies.}
\label{tab:appdata}
\begin{tabular*}{\textwidth}{llp{9.8cm}}
\toprule%
& \textbf{Size} & \textbf{Dataset Description} 
\\ \toprule
& 
11,937 Posts &
The Bridging Benchmark comments dataset \cite{schmer2024annotator} annotated for the presence of alienation, compassion, curiosity, moral outrage, reasoning, and respect.
\\ \midrule
& 
3,718 Posts &
Posts sampled from Ribeiro et al.'s (\citeyear{ribeiro2021evolution}) analysis of the ``Manosphere.''
\\ \midrule
\multirow{4}{*}{\rotatebox[origin=c]{90}{In MFTC \cite{hoover2020moral}}} &
4,424 Tweets &
MFTC ``ALM:'' related to the All Lives Matter movement.
\\ \cmidrule{2-3}
&
5,480 Tweets &
MFTC ``Baltimore:'' related to the Baltimore protests.
\\ \cmidrule{2-3}
&
4,847 Tweets &
MFTC ``Davidson'' from Davidson's \cite{davidson2017automated} hate speech corpus.
\\ \cmidrule{2-3}
&
4,890 Tweets &
MFTC ``MeToo:'' tweets related to the \#MeToo movement.
\\ \botrule
\end{tabular*}
\end{minipage}
\end{table}

Our application studies are intended to be simple, focused analyses to confirm social science hypotheses and explore directed regard within online communities.
In all of these studies, we processed the various texts in the dataset with the same DSR models and the same initialization parameters.
We did not use any domain-specific or dataset-specific lexicons or settings.
After the datasets were processed by DSR, we employ different statistical tests and visualizations to that show a mix of intuitive and counter-intuitive findings via DSR.

We first demonstrate DSR analyses that relate intensity scores (e.g., Oppose–Advocate) to other, pre-annotated attributes (e.g., Moral Outrage and Respect), and we show how DSR can accumulate these by density or by category (e.g., first-person, second-person, and genders) in \secref{app-bridging}.

\subsubsection{The Bridging Benchmark Dataset}
\label{sec:app-bridging}

The DSR problem formulation and dataset are specified at the \emph{span} or \emph{phrase}-level (\secref{data}), but the Bridging Benchmark dataset \cite{schmer2024annotator} and Moral Foundations Twitter Corpus \cite{hoover2020moral} are labeled at the \emph{document}-level.
To relate these different levels of detail, we can assess how various document-level labels---in this case, \emph{Moral Outrage} and \emph{Respect} from the Bridging Benchmark---are related to the DSR scores of the corresponding spans in those documents.

We processed all 11,937 distinct PAI texts from the open-source Bridging Benchmark dataset with DSR to extract and score all topic and character spans per document.
We then segment the results into \emph{High} and \emph{Low} bins for each attribute based on annotator agreement in the Bridging Benchmark data: High indicates $\geq$ 4/5 raters marked it positive; Low indicates that $\geq$ 4/5 raters marked it negative.

\begin{figure}[htb]
\centering
\includegraphics[width=\textwidth]{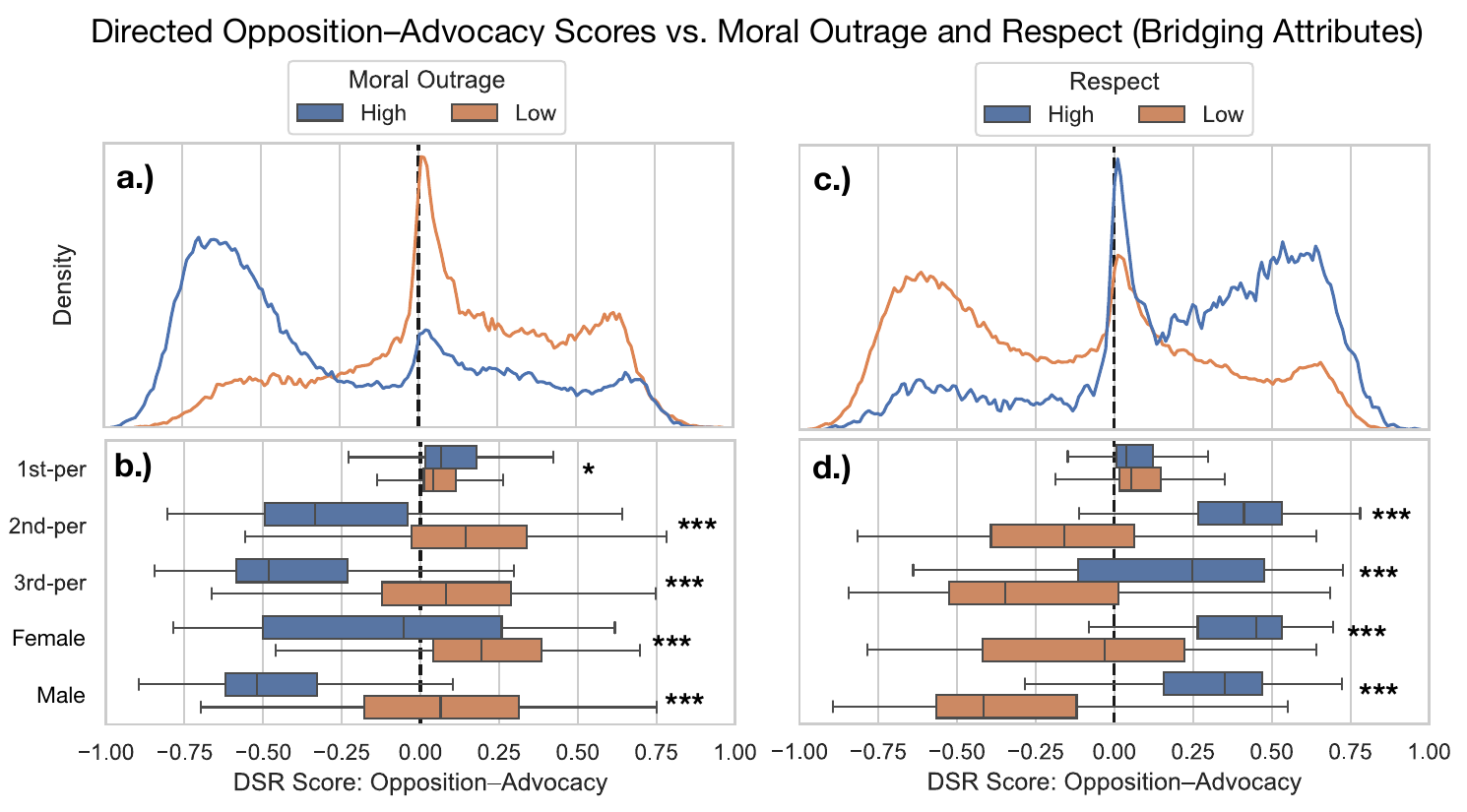}
\caption{Document-level Moral Outrage (left) and Respect (right) labels from the Bridging Benchmark plotted against span-level Opposition–Advocacy scores.  Moral outrage is associated with more intense opposition toward 2nd and 3rd-person characters, but more intense advocacy toward 1st-person ($b$).  Respect is associated with more intense advocacy toward 2nd and 3rd-person characters, with no effect on 1st-person regard ($d$).
(***: $p<.0001$, **: $p<.001$; *: $p<.01$; $\sim$: $p<.05$.)}
\label{fig:bridging}
\end{figure}

\figref{bridging} shows the results of this analysis for two representative Bridging Benchmark attributes: \emph{Moral Outrage} in \figref{bridging}(a-b) and \emph{Respect} in \figref{bridging}(c-d).
As shown in the density plot in \figref{bridging}($a$), spans within documents with \emph{High} Moral Outrage contribute to a peak of opposition ($x < 0$), and spans within documents with \emph{Low} Moral Outrage contribute to a peak of neutrality ($x = 0$) and more frequent advocacy ($x > 0$).
High Moral Outrage yielded significantly lower scores of Opposition–Advocacy ($M=-.25, SD=.46$) than Low Moral Outrage $t(1998)=-21.46, p<0.0001$
The converse is true of \emph{High} and \emph{Low} Respect, as shown in \figref{bridging}($c$), where High Respect yielded significantly higher Opposition–Advocacy scores ($M=.23, SD=.39$) than Low Respect $t(1998)=21.54, p<0.0001$.

Since DSR surfaces span-based results from raw text, we can also aggregate the results based on the subject of regard, or at whom or what the regard is \emph{directed}.
We use a domain-independent listing of pronouns (see \tabref{pronouns} in the Appendix) and simple male/female nouns to aggregate character spans into 1st person, 2nd person, 3rd person, and female- and male-specific nouns/pronouns.
These results are plotted in \figref{bridging} ($b$) and ($d$) for Moral Outrage and Respect, respectively.
In keeping with their respective density plots, the 2nd-person, 3rd-person, Female, and Male referring spans all score significantly lower ($p < 0.0001$) Opposition–Advocacy for High Moral Outrage, and these categories all score significantly higher for High Respect.

Notably, in High Moral Outrage, the first-person references (e.g., I, me, we, us, etc.) had slightly higher advocacy ($M=.1, SD=.15$) than Low Moral Outrage $t(1998)=2.52, p<0.01$, which is the opposite pattern of the other categories, and the High and Low Respect conditions had no significant effect on first-person advocacy.
These results suggest that most instances of Moral Outrage and Respect in the Bridging Benchmark are directed at others in the community, and less self-directed toward the author.
The increased self-advocacy in Moral Outrage may indicate that authors sometimes use an in- and out-group dichotomy, but the effect is small.


The DSR results presented here suggest that DSR does not obviate or replace these Bridging attributes (which are motivated by different social science theories); however, DSR correlates intuitively with these document-level attributes as an complementary analysis.

\subsubsection{The Moral Foundataions Twitter Corpus}
\label{sec:app-mftc}

The Moral Foundataions Twitter Corpus (MFTC) datasets (\tabref{appdata}) include human annotations of Moral Foundations (MF) attributes, which are organized into \emph{virtue-vice} pairs according to Moral Foundations theory: \textit{care-harm}, \textit{fairness-cheating}, \textit{loyalty-betrayal}, \textit{authority-subversion}, and \textit{purity-degradation}. 
These attributes are intuitively named, but for details see Hoover et al. \cite{hoover2020moral}.

We hypothesize that the occurrence of machine-predicted DSR attributes will correlate meaningfully with the human-labeled MFTC virtues and vices.
In this analysis, we use the MF attribute labels on each tweet from the MFTC dataset as independent variables, and we use the span-level DSR scores from those messages as dependent variables.
We use a thresholding approach to convert the numerical DSR intensity scores into categorical labels.
Since the DSR scores are predicted along three dimensions, as shown in \figref{dsr-ui}, we can use a threshold value $\sigma$ on each of the three dimensions, e.g., to compute that \figref{dsr-ui}'s \textq{they} and \textq{do this} are both Opposed (\textbf{oppo/advo} $\leq -\sigma$) and Harmful (\textbf{harmf/helpf} $\leq -\sigma$), and that \textq{children} are Advocated (\textbf{oppo/advo} $\geq \sigma$) and Victimized (\textbf{victm/aided} $\leq -\sigma$).
For this and the other studies, we use $\sigma = 0.15$ to capture mild and moderate occurrences of these attributes.

Converting from intensity scores to categorical labels allows us to perform a frequency analysis, e.g., to assess the relative occurrence of these DSR attributes occurring alongside MF attributes.
A single MFTC-labeled tweet often contains many character and topic spans that are predicted and scored by DSR, so this analysis measures the log odds ratio that a span has a given DSR label (e.g., Oppose, Advocate, etc.), provided it is from a tweet with a given MF label.
For these statistical tests we use Fisher's Exact Test in SciPy to compute the odds ratio and the exact $p$-value and then we report the log odds ratio for display purposes.

\begin{figure}[htb]
\centering
\includegraphics[width=\textwidth]{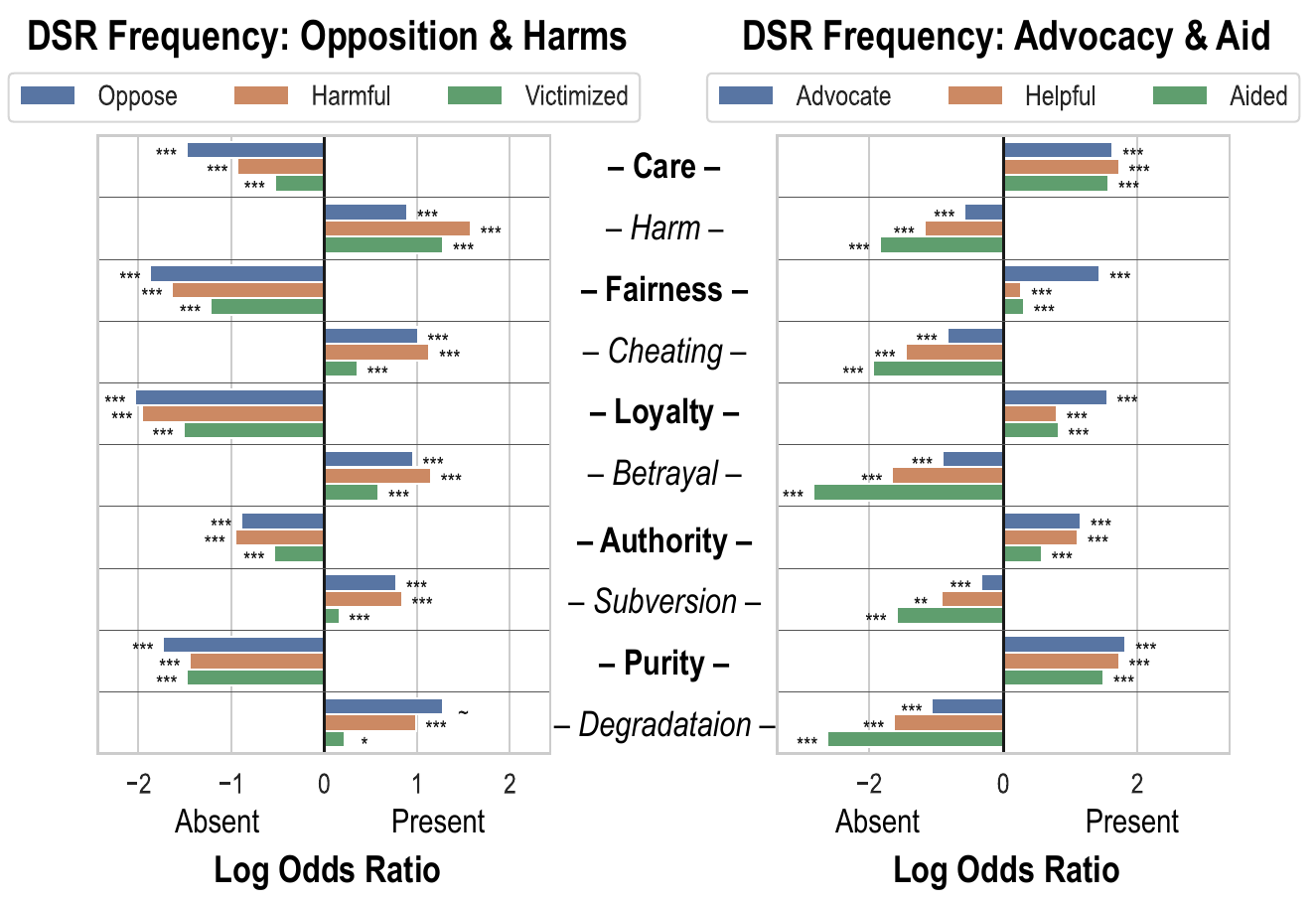}
\caption{Log odds ratio of machine-predicted DSR attribute occurrences for each human-assigned MF label in the MFTC dataset.
At left, the negative poles of the DSR dimensions (Oppose, Harmful, and Victimized) are plotted for each MFTC virtue (bold) and vice (italics); at right, the corresponding positive poles of the DSR dimensions (Advocate, Helpful, Aided) are plotted for the same MFTC attributes (***: $p<.0001$, **: $p<.001$; *: $p<.01$; $\sim$: $p<.05$.).}
\label{fig:mftc}
\end{figure}

\figref{mftc} plots the results of our DSR analysis against MFTC attributes:
the left chart plots the negative poles of the three DSR dimensions (Oppose, Harmful, and Victimized) and
the right chart plots the respective positive poles (Advocate, Helpful, and Aided).
All MF virtues (bold labels in the \figref{mftc} center column) are associated with significantly lower likelihood of Oppose, Harmful, and Victimized DSR predictions, and significantly higher likelihood of Advocate, Helpful, and Aided DSR predictions.
Conversely, all MF vices (italicized labels in the \figref{mftc} center column) are associated with higher likelihood of Oppose, Harmful, and Victimized DSR predictions, and lower likelihood of Advocate, Helpful, and Aided.
Intuitively, we observe the Harmful DSR spans most frequently in MFTC tweets labeled with Harm, and we observe Aided and Helpful DSR spans with highest frequency in MFTC tweets labeled with Care.

As with our analysis of DSR against the Bridging Benchmark dataset above, this MFTC analysis shows that attributes focused on other social science theories (e.g., moral factors) correlate meaningfully with the DSR attributes concerning directed blame/helpfulness, directed victimization/aid, and directed opposition/advocacy.
Furthermore, it shows we can meaningfully correlate attributes at the document-level (or tweet-level) with DSR attributes at the span-level.

\subsubsection{Comparing corpora for composite patterns of regard}
\label{sec:app-log-odds}
\label{sec:app-composite}

The above studies characterized how span-level regard co-occurs with document-level attributes from other social science theories.
The final three application studies demonstrate the use of DSR for comparative corpus analysis, agnostic to the theories of Moral Foundations and of bridging attributes.

In this study we compare the four MFTC corpora and the Ribeiro corpus (all described in \tabref{appdata}) using their log odds ratios of composite patterns of regard.
Rather than characterizing the frequency of singular spans scored as Advocated or Harmful, we test joint attributes, e.g., spans that are scored as \emph{both} Advocated and Harmful, where the author advocates for a harmful act or actor.
We also test conditional attributes, e.g., Victim|Female, 
which is the frequency of a span being scored as Victimized given that it is a Female noun or pronoun.

\begin{figure}[htb]
\centering
\includegraphics[width=\textwidth]{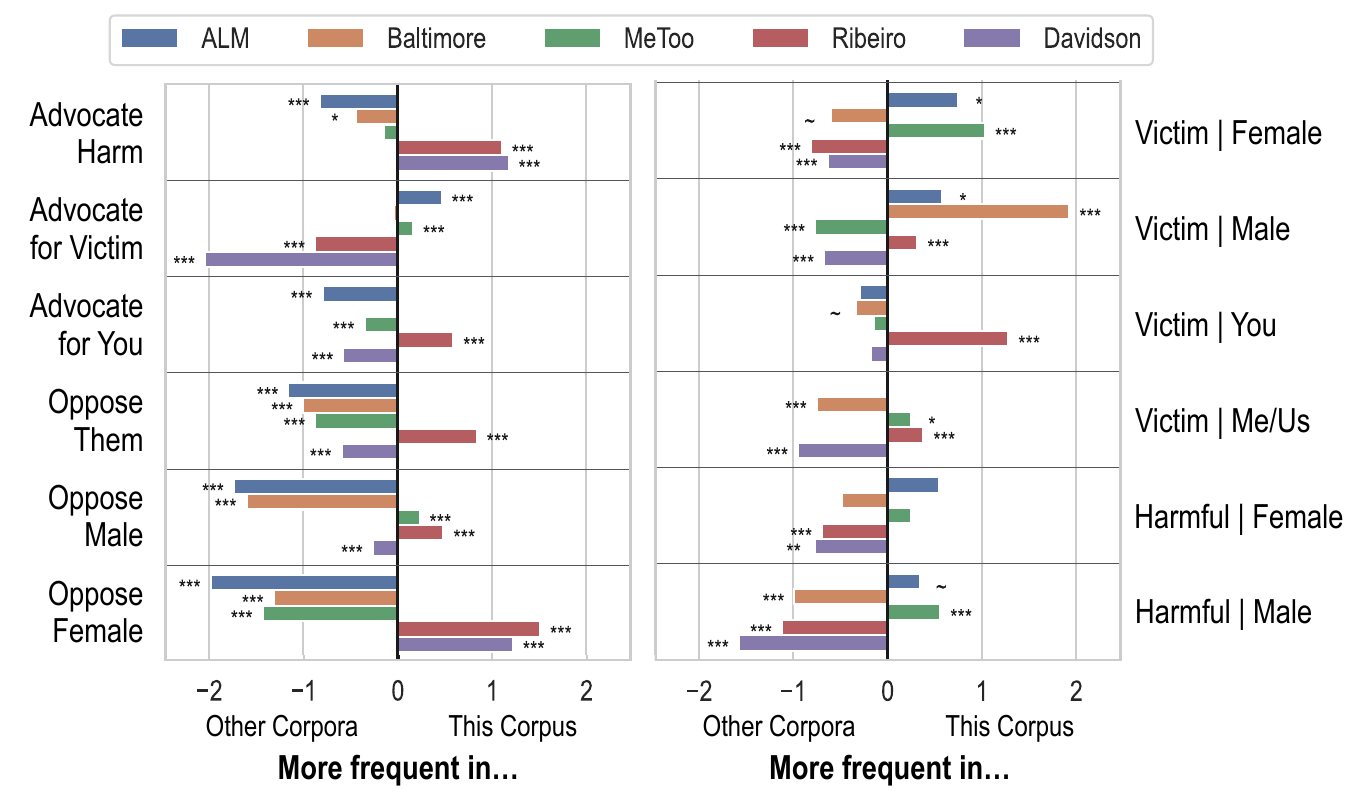}
\caption{Comparison of five different corpora using joint attributes (left) and conditional attributes (right).
Both plots report log odds ratios across levels of statistical significance (***: $p<.0001$, **: $p<.001$; *: $p<.01$; $\sim$: $p<.05$.).}
\label{fig:composite}
\end{figure}

Results of this analysis are plotted in \figref{composite}, with joint attributes at left and conditional attributes at right, reporting log odds ratio of the attribute in the given corpus relative to all other corpora.
Joint attributes measure the co-occurrence of two labels: ``Advocate for Victim'' (\figref{composite}, left) is the occurrence of both ``Advocate'' and ``Victimized'' on the same label.
Conditional attributes are the occurrence of the first label given the second, e.g., ``Victim$|$You'' (\figref{composite}, right) is the occurrence of ``Victimized'' provided the span is a second-person span such as ``you,'' ``your,'' ``yourselves,'' etc.

The joint attribute ``Advocate Harm'' in \figref{composite} (top left) is one of the most pernicious, and we observe significantly higher frequencies of this in the Davidson hate speech corpus and the Ribeiro manosphere corpus.
Advocacy for harm frequently manifests as threats in hate speech, and often as retaliation against feminism in the Ribeiro corpus.
These two datasets also exhibit significantly lower ``Advocate for Victim'' than other corpora, and both exhibit significantly higher ``Oppose Female.''
The opposition to female nouns and pronouns is expected in manosphere content (often opposing feminist ideals) and also in hate speech (often disrespecting women).

Additional corpus-relevant findings are evident in the \figref{composite} findings.
Many of the Ribeiro manosphere messages are responding to a forum member (i.e., ``you'') that they advocate for (highest ``Advocate for You,'' $p<.0001$) and characterize as victimized (highest ``Victim$|$You,'' $p<.0001$), and they also characterize themselves as victimized (i.e., highest ``Victim$|$Me/Us,'' $p<.0001$), e.g., by feminist ideologies that they oppose, often cast as an antagonistic ``they'' or ``them'' (highest ``Oppose Them,'' $p<.0001$).
These four manosphere findings are not observed in the MFTC Davidson hate speech corpus, which has significantly lower ``Advocate for You,'' lower ``Oppose Them,'' and lowest ``Victim$|$Me/Us.''

The MFTC MeToo corpus has the significantly highest incidence of ``Victim$|$Female'' ($p<.0001$) and ``Harmful$|$Male'' ($p<.0001$), both consistent with the themes of the \#MeToo movement.

The MFTC ALM and Baltimore corpora both contain messages about violence and law enforcement.
Both have significantly higher ``Advocate for Victim'' and lower ``Advocate for Harm,'' though the Baltimore corpus focuses primarily on a male victim of violence (highest ``Victim$|$Male,'' $p<.0001$), and ALM does not make this distinction. 
Our next DSR analysis will further contrast the MFTC ALM and Baltimore corpora, among others.

This composite attribute results in \figref{composite} plots twelve representative attributes that combine DSR indicators (e.g., ``Advocate,'' ``Oppose,'' ``Harmful,'' ``Victim,'' etc.) and span-level identifiers (e.g., ``You,'' ``Them,'' ``Me/Us,'' ``Female,'' and ``Male'') to show that we can meaningfully analyze the \emph{target} of directed regard and distinguish gender-related or ``us''-versus-``them'' related patterns of regard.
Many additional composite indicators can be crafted similar to these, e.g., for geopolitical or sociocultural concepts in addition to the pronoun-oriented concepts demonstrated here.

\subsubsection{Characterizing corpora by the target and intensity of regard}
\label{sec:app-diff}

\begin{figure}[htb]
\centering
\includegraphics[width=\textwidth]{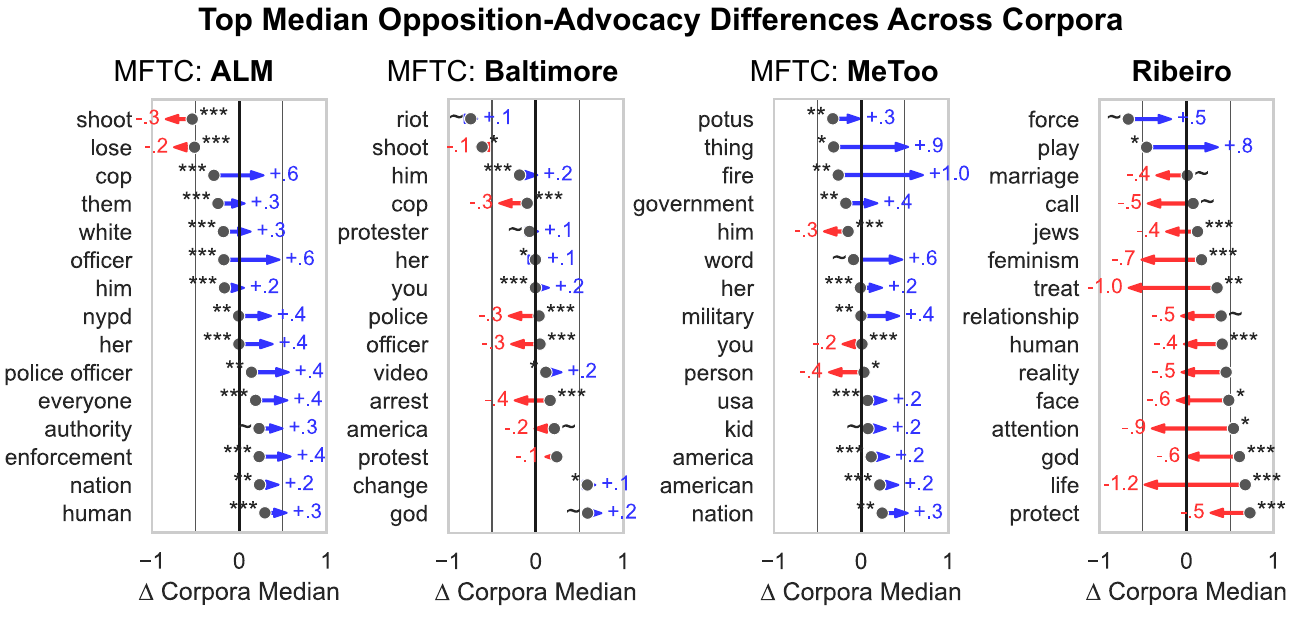}
\caption{Three MFTC corpora \cite{hoover2020moral} and the Ribeiro corpus \cite{ribeiro2021evolution} are differentiated by the top 15 highest-delta objects of regard against all other corpora.  Gray points are other corpora's median regard, and labeled blue (red) arrows plot the increased (decreased) median opposition-advocacy in the plotted corpus (***: $p<.0001$, **: $p<.001$; *: $p<.01$; $\sim$: $p<.05$.).}
\label{fig:diff}
\end{figure}

A primary capability of DSR is surfacing the target of regard, e.g., by identifying the span ``children'' as a central target of regard in the \figref{dsr-ui} example ``How can they do this to the children??'' and predicting intense advocacy and victimization to ``children.''
This section exemplifies a target-centric analysis: we accrue all targets of regard that have at least 20 occurrences within a corpus and at least 20 combined occurrences in other corpora, and then we compare the in-corpus and out-of-corpus distributions of Oppose–Advocate scores for those topics.
Results are plotted in \figref{diff}, comparing three MFTC corpora (ALM, Baltimore, and MeToo) with the Ribeiro corpus.

\figref{diff} lists the top 15 largest differences in median Oppose–Advocate intensity for each corpus, compared to all other corpora, where $x=-1$ is intense Oppose, $x=0$ is neutral, and $x=1$ is intense Advocate.
The gray dots show the median intensity of other corpora, and the arrows point to the median intensity of the corpus in question, with asterisks indicating significance of t-test (calculated in SciPy) between corpus intensities and other corpora intensities.

Results in \figref{diff} show differences in regard between ALM and Baltimore MFTC corpora: both corpora frequently mention concepts surrounding law enforcement and authority, but in ALM we observe significantly higher Advocacy directed at these concepts, and in Baltimore we observe significantly higher Opposition to the same concepts.
For instance, ``cop'' and ``officer'' are both scored $+.6$ above the corpora-complement median in ALM and both are scored $-.3$ below the corpora-complement median in Baltimore.
This exemplifies how DSR can surface opposing topic-directed regard across corpora from the bottom-up, without having to specify ``officer,'' ``police,'' or ``law enforcement'' \textit{a priori} in our analysis. 

The MeToo MFTC corpus includes some positive sentiment toward the U.S. government and military and also statistically significant gender-based advocacy, corroborating the frequency analysis in \secref{app-log-odds}.

For the Ribeiro manosphere corpus, all but two of the top median differences from other corpora are in the Oppose direction.
The most pronounced difference is the $-1.2$ median Opposition of ``life'' compared to other corpora.
Our spot-checks of this reveal that in this corpus, most mentions of ``life'' are in contexts such as ``I hate my life,'' ``Life is... cruel,'' and other lamentations.

This analysis characterizes both the object of regard (e.g., ``life'') and the difference with other corpora or a baseline (e.g., $-1.2$).
This surfaces topics and characters that are regarded in an extreme or distinctive fashion by an audience or corpus. 
As with our previous analyses, this is all based on the same DSR NLP output that identifies textual spans and scores them with numerical intensity.

\subsubsection{Visualizing pairwise patterns of regard}
\label{sec:app-pairwise}

The above analyses focus on single targets of regard, e.g., the corpus-level advocacy or opposition toward law enforcement; however, many themes in online messaging have more structure, e.g., the message is not simply that ``we'' are victimized, but rather ``we'' are victimized by ``them.''
We can use DSR to identify frequently-occurring pairs of entities in a corpus such that one is often scored as Victimized when another is scored as Harmful, or one is often scored as Helpful when another is scored as Aided.
These topic-pairs comprise micro-narratives in a corpus, and we can visualize these pairs using a knowledge graph.

To construct our pairwise view, we use the same threshold value ($\sigma=0.15$) as in \secref{app-mftc} and we collect pairs of entities within messages for two patterns: (1) \textbf{harm}($A,B$) where \textit{harmful\_helpful}($A$) $\leq -\sigma$ and \textit{victimized\_aided}($B$) $\leq -\sigma$; and (2) \textbf{help}($A,B$) where \textit{harmful\_helpful}($A$) $\geq \sigma$ and \textit{victimized\_aided}($B$) $\geq \sigma$.
Note that \textbf{harm}($A,B$) and \textbf{help}($A,B$) do not capture indicate that $A$ \emph{directly} harms or helps $B$; rather, these patterns indicate a persistent micro-narrative in the messaging such that $A$ is frequently harmful (or helpful) when $B$ is victimized (or aided).

\begin{figure}[htb]
\centering
\includegraphics[width=.8\textwidth]{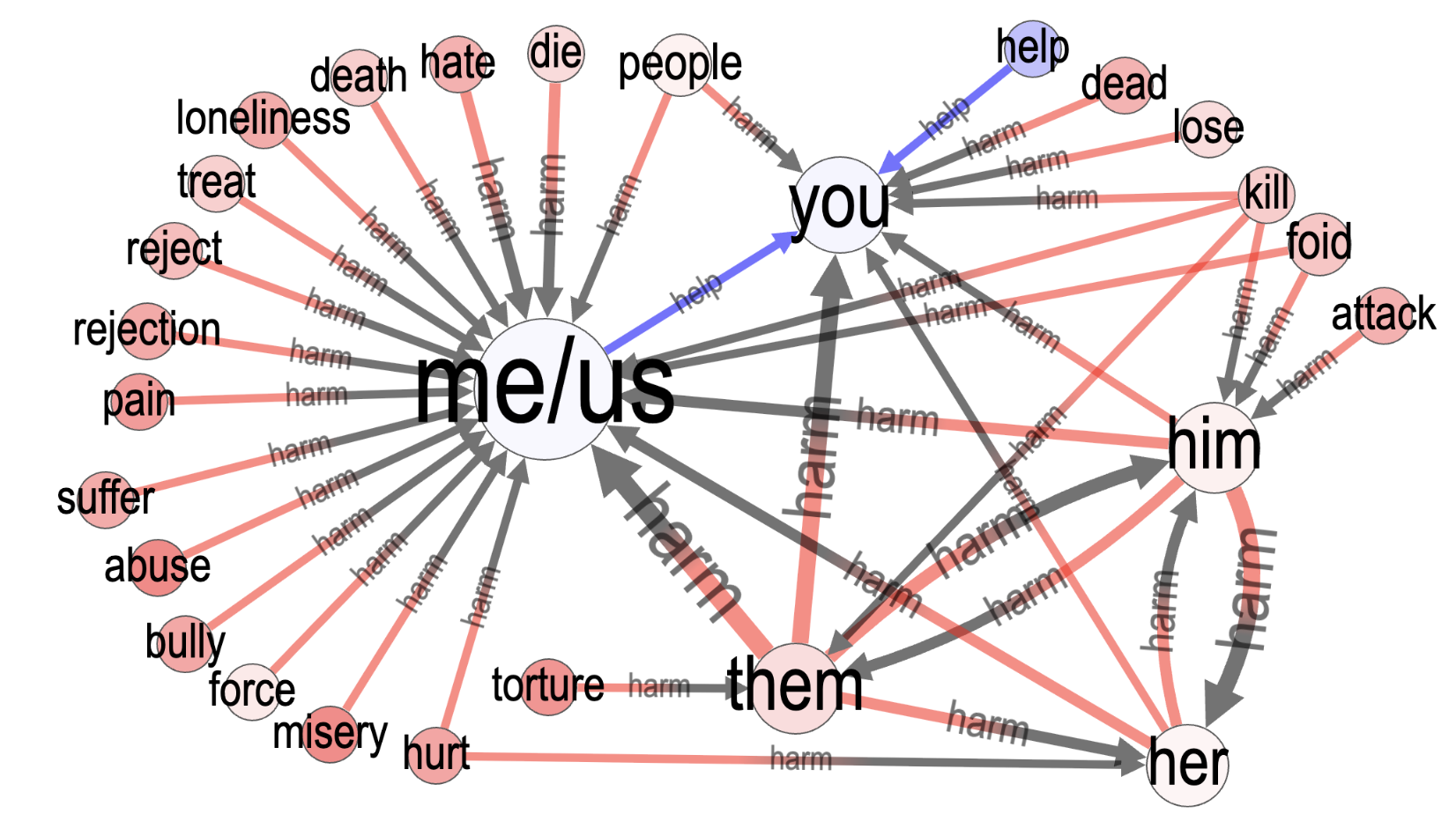}
\caption{Pairwise theme plot of the Ribeiro manosphere corpus.}
\label{fig:ribeiro-pairwise}
\end{figure}

\begin{figure}[htb]
\centering
\includegraphics[width=.8\textwidth]{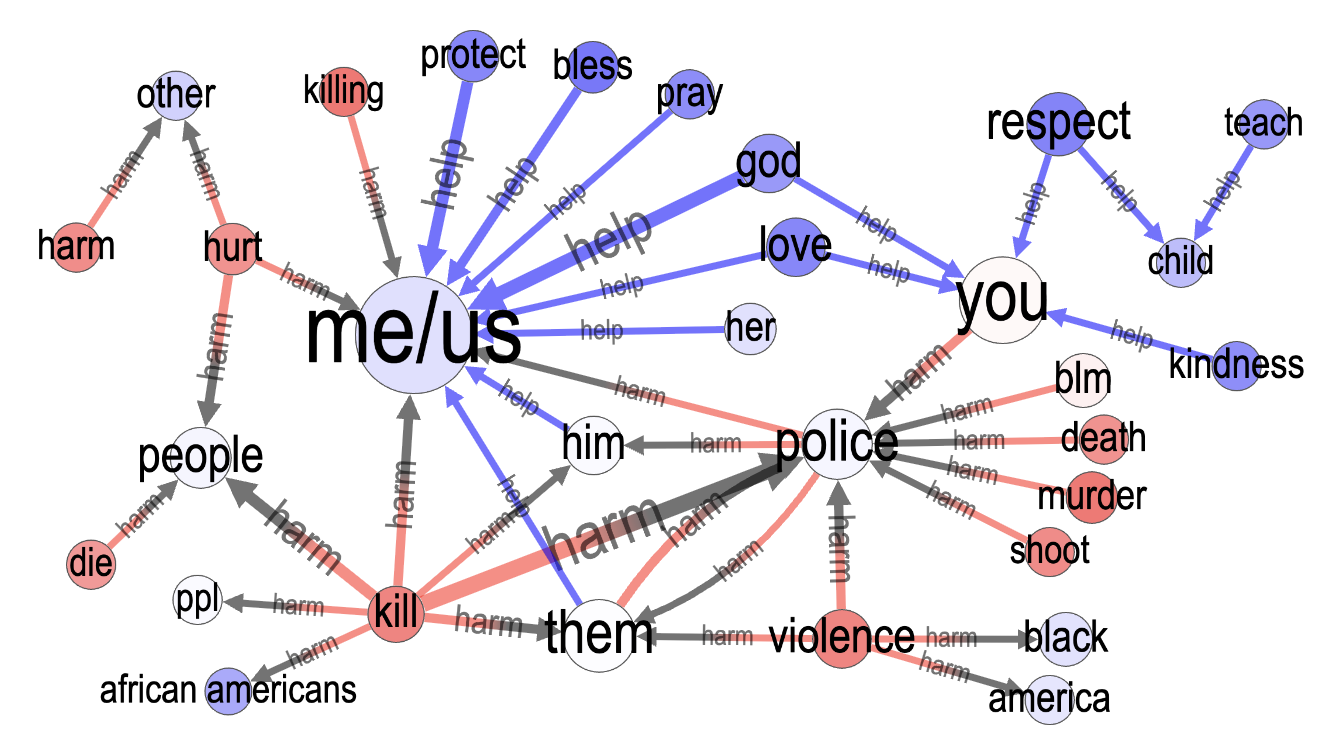}
\caption{Pairwise theme plot of the MFTC ALM corpus.}
\label{fig:alm-pairwise}
\end{figure}

There are thousands of distinct \textbf{harm} and \textbf{help} pairs in each corpus, so we visualize the 40 most frequent pairs for simplicity.
To demonstrate this capability we extract and display pairwise regard for the Ribeiro corpus (\figref{ribeiro-pairwise}) and the MFTC ALM corpus (\figref{alm-pairwise}).
In each, we scale the nodes and edges by the frequency of occurrence in the corpus (bigger is more frequent), we color the nodes by their corpus level opposition–advocacy (red is more opposed, blue is more advocated, white is neutral), and we label and color the edges based on their \textbf{harm} or \textbf{help} pattern.
The resulting visualization helps express the frequency of the various pairwise themes and also the general advocacy and opposition levied on each constituent topic.

The pairs for the Ribeiro corpus in \figref{ribeiro-pairwise} center around a ``me/us'' that is harmed by 20 various events and actors, such that half of the top 40 pairs express \textbf{harm} to ``me/us.''
The most frequent pair is \textbf{harm}(them, me/us), and of the 40 top pairs, there are only two \textbf{help} pairs, both aiding ``you.''
The frequency and multitude of terms for harms to the first-person (both singular and plural) are consistent with the general self-victimization themes in this manosphere forum.
Some other representative themes are the cyclic \textbf{harm}(him, her) and \textbf{harm}(her, him), indicating discussions of conflict across genders, and the presence of the harmful and opposed ``foid'' with \textbf{harm} to ``me/us'' and ``him.''
In this forum, ``foid'' (short for ``feminoid'') is a dehumanizing term for girls and women, and it is not often used outside of the manosphere.

The 40 most frequent pairwise themes extracted from the MFTC ALM corpus are shown in \figref{alm-pairwise}.
These include a higher share of \textbf{help} pairs (14/40) than the Ribeiro manosphere corpus, including high advocacy for religious terms (``bless,'' ``pray,'' and ``god,'') all related to the aided ``me/us.''
The most frequent victimized character within this chart is ``police,'' where eight of the 40 most frequent pairs express \textbf{harm} to this target.

These pairwise themes help characterize frequent micro-narratives in each corpus, ranging from broad self-victimization, inter-gender conflict, commitment to faith, and harms to law enforcement.
These themes are consistent with Ribeiro's analysis of the manosphere \cite{ribeiro2021evolution} and of the MFTC selection of the ALM corpus \cite{hoover2020moral}, so the automated DSR findings presented here corroborate the human analyses or Ribeiro et al., and the data collection criteria of the MFTC curators.

As with most knowledge graph extractions and visualizations, these pairwise analyses do not afford statistical observations like the above analyses.
Consequently, we regard pairwise themes as a corpus-mining capability that is only as accurate as the DSR spans and scores they are inferred from.
This pairwise analysis is motivated in part by social science theories that acknowledge the combination of sentiments, regards such as compassion for victims triggering anger toward their aggressors \cite{graham2013moral}.

We detected and visualized two pairwise theme labels (\textbf{harm} and \textbf{help}), but other pairs are possible, e.g., finding symmetric co-harmful topics, co-victimized topics, and so-forth.

%% file: discussion.tex
\section{Discussion and Conclusions}
\label{sec:discussion}

This paper presents the directed social regard (DSR) problem setting with social science motivation (\secref{socialscience}), data annotation methodology (\secref{data}), dual ML/NLP models (\secref{model}), validation studies (token classification in \secref{eval-ner} and numerical sentiment analysis in \secref{eval-absa}) and five applications on pre-existing datasets in \secref{application}.

The DSR problem setting (\secref{socialscience}) characterizes nuanced expressions of opposition–advocacy, harmful–helpful, and victimized–aided directed at specific span-level characters and topics.
These span-level DSR dimensions are related to document-level moral foundations attributes \cite{hoover2020moral} and bridging attributes \cite{schmer2024annotator} from prior work; however, unlike these document-level attributes, the DSR span-level attributes characterize prosocial (advocacy, helpfulness, aid) and antisocial attributes (opposition, harmfulness, victimization) within the same sentence and targeted at different topics.

Our analyses of the Bridging Benchmark and Moral Foundations datasets show that DSR  scores correlate meaningfully with previous message-level bridging attributes (\secref{app-bridging}), and that DSR attribute frequency correlates meaningfully with message-level moral foundations attributes (\secref{app-mftc}).

%% file: appendix.tex
\section{Appendix*}


\begin{table}[h]
\begin{minipage}{\linewidth}
\caption{Referring phrases used across datasets to categorize Character spans into groups.}
\label{tab:pronouns}
\begin{tabular*}{\textwidth}{lp{9cm}}
\toprule%
\textbf{Category (Display)} & \textbf{Lemmas} 
\\ \midrule
1st-Person (me/us) &
I, me, my, mine, myself, ourself, ourselves, our, ours, we, us
\\ \midrule
2nd-Person (you) &
you, your, yours, yourself, yourselves, u, ur
\\ \midrule
3rd-Person-Female (her) &
she, her, hers, herself, female, woman, girl, lady
\\ \midrule
3rd-Person-Male (him) &
he, him, his, himself, man, boy, guy, male
\\ \midrule
3rd-Person-Misc (them) &
they, them, their, theirs, themselves, themself, those
\\ \botrule
\end{tabular*}
\end{minipage}
\end{table}